\documentclass[letterpaper]{article} % DO NOT CHANGE THIS
\usepackage{aaai2026}  % DO NOT CHANGE THIS
\usepackage{times}  % DO NOT CHANGE THIS
\usepackage{helvet}  % DO NOT CHANGE THIS
\usepackage{courier}  % DO NOT CHANGE THIS
\usepackage[hyphens]{url}  % DO NOT CHANGE THIS
\usepackage{graphicx} % DO NOT CHANGE THIS
\usepackage{natbib}  % DO NOT CHANGE THIS AND DO NOT ADD ANY OPTIONS TO IT
\usepackage{caption} % DO NOT CHANGE THIS AND DO NOT ADD ANY OPTIONS TO IT
\usepackage{algorithm}
\usepackage{algorithmic}
\usepackage{amssymb}
\usepackage{amsmath}
\usepackage{booktabs}
\usepackage{multirow}
\usepackage{newfloat}
\usepackage{listings}
\DeclareCaptionStyle{ruled}{labelfont=normalfont,labelsep=colon,strut=off} % DO NOT CHANGE THIS
\floatstyle{ruled}
\newfloat{listing}{tb}{lst}{}
\floatname{listing}{Listing}
\title{ReCast: Reliability-aware Codebook Assisted Lightweight Time Series Forecasting}
\author{
Xiang Ma\textsuperscript{\rm 1}, 
Taihua Chen\textsuperscript{\rm 1,2}, 
Pengcheng Wang\textsuperscript{\rm 1}, 
Xuemei Li\textsuperscript{\rm 1}, 
Caiming Zhang\textsuperscript{\rm 1}
}
\affiliations{
\textsuperscript{\rm 1}Shandong University, Jinan, China\\
\textsuperscript{\rm 2}Joint SDU-NTU Centre for Artificial Intelligence Research (C-FAIR), Shandong
University, Jinan, China\\

\{xiangma,xmli,czhang\}@sdu.edu.cn, \{cfair-cth,pengchengwang\}@mail.sdu.edu.cn
}

\begin{document}

\maketitle

\begin{abstract}
	Time series forecasting is crucial for applications in various domains. Conventional methods often rely on global decomposition into trend, seasonal, and residual components, which become ineffective for real-world series dominated by local, complex, and highly dynamic patterns. Moreover, the high model complexity of such approaches limits their applicability in real-time or resource-constrained environments. In this work, we propose a novel \textbf{RE}liability-aware \textbf{C}odebook-\textbf{AS}sisted \textbf{T}ime series forecasting framework (\textbf{ReCast}) that enables lightweight and robust prediction by exploiting recurring local shapes. ReCast encodes local patterns into discrete embeddings through patch-wise quantization using a learnable codebook, thereby compactly capturing stable regular structures. To compensate for residual variations not preserved by quantization, ReCast employs a dual-path architecture comprising a quantization path for efficient modeling of regular structures and a residual path for reconstructing irregular fluctuations. A central contribution of ReCast is a reliability-aware codebook update strategy, which incrementally refines the codebook via weighted corrections. These correction weights are derived by fusing multiple reliability factors from complementary perspectives by a distributionally robust optimization (DRO) scheme, ensuring adaptability to non-stationarity and robustness to distribution shifts. Extensive experiments demonstrate that ReCast outperforms state-of-the-art (SOTA) models in accuracy, efficiency, and adaptability to distribution shifts.
\end{abstract}

%\begin{links}
%    \link{Extended version}{https://aaai.org/example/extended-version}
%\end{links}

\section{Introduction}
In recent years, time series forecasting has gained significant attention due to its critical applications in various real-world applications, including finance, energy, healthcare, and industrial automation \cite{wen2023transformers,ma2023dynamic,qiu2024tfb,shibo2025hdt}. Capturing complex and irregular temporal patterns accurately remains a primary challenge in this domain. Conventional approaches typically address this complexity by globally decomposing time series into trend, seasonal, and residual components, and modeling these components independently \cite{Autoformer,Fedformer,hu2025adaptive}. However, while effective for structured or periodic data, such global decomposition methods often underperform when faced with dynamic and noisy real-world time series \cite{tang2025unlocking}. Moreover, these methods typically involve considerable model complexity, which limits their practicality in resource-limited environments \cite{ansarichronos}.

To address these challenges, we introduce a novel \textbf{RE}liability-aware \textbf{C}odebook-\textbf{AS}sisted \textbf{T}ime series forecasting (ReCast) framework, focusing explicitly on capturing local patterns. Observing that many real-world series exhibit recurring local shapes rather than clear global regularities \cite{yeh2016matrix}, ReCast quantizes these local shapes into a learnable codebook, generating discrete embeddings to represent evolving patterns. This codebook-based representation not only captures salient local structures but also reduces model complexity, enabling a inherently lightweight forecasting design. Meanwhile, residual modeling is introduced to capture irregular variations not adequately represented by the quantized embeddings, ensuring robustness to fluctuations without excessively increasing model size. 

Specifically, ReCast segments input into patches, quantifying each as discrete embedding using a dynamically updated reliability-aware codebook. As shown in Figure \ref{framework}, a quantization path is used to forecast the future discrete embeddings, and a residual path learns to estimate the difference between input and its approximate representation reconstructed by discrete embedding. These two paths work in synergy: the quantization path enables lightweight forecasting of regular structures, while the residual path ensures the reliable reconstruction of irregular fluctuations. The prediction results combine outputs from both paths. To reduce overfitting and improve generalization to distribution shifts, we perform random patch sampling, and select only a subset of patches for training and codebook updates. Downsampling is applied prior to quantization, helping to highlight salient local structures and lower computational cost. 

More importantly, it can be observed that the performance of ReCast strongly depends on the stability and adaptability of the codebook \cite{guo2023compact}. Therefore, we propose an incremental codebook update mechanism centered on a reliability-aware scoring method. This method can robustly guide the update process in response to evolving data distributions, striking a balance between stability and adaptability. Our \textbf{contributions} include:
\begin{figure*}
	\centering
	\includegraphics[width=\linewidth]{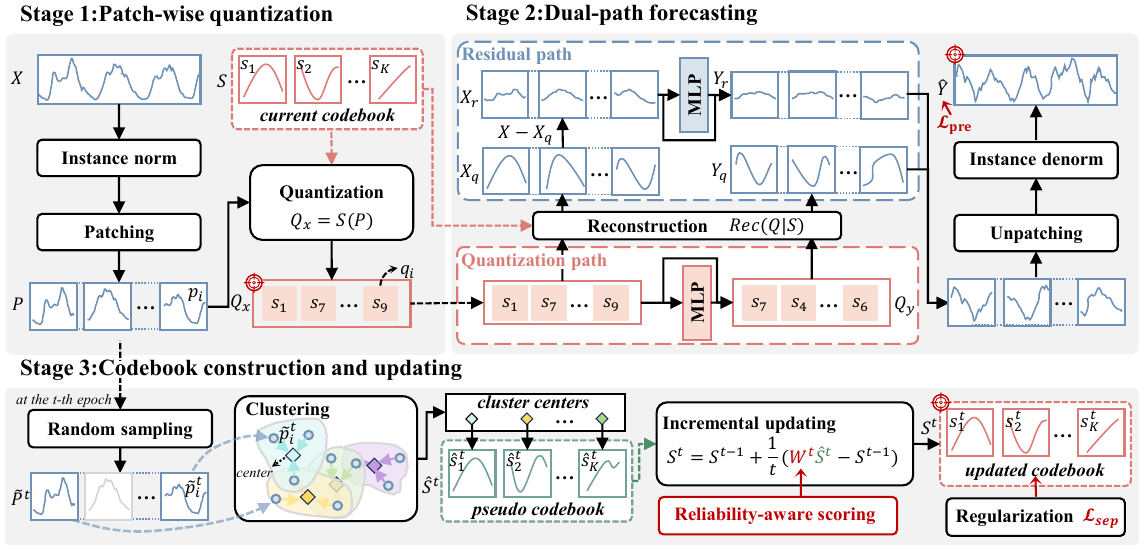}
	\caption{ReCast overview. It comprises patch-wise quantization, dual-path forecasting, and codebook construction and updating.}
	\label{framework}
\end{figure*}
\begin{itemize}
	\item We propose a codebook assisted lightweight forecasting framework that effectively captures both regular and irregular local temporal patterns while significantly reducing model complexity.
	
	\item We introduce a reliability-aware updating mechanism for codebook, which enhances adaptability and robustness to noise and distribution shifts with low computational cost.
	
	\item Extensive experiments show that ReCast achieves superior accuracy, generalization, and robustness, relying on its lightweight architecture and efficient training strategy.
\end{itemize}
\section{Related Work}
\subsection{Deep Learning-based Forecasting}
Recent advances in deep learning have significantly improved time series forecasting by leveraging powerful representation learning capabilities. The convolutional neural network(CNN)-based approaches \cite{wu2023timesnet} introduce local receptive fields to capture short-term dynamics efficiently, but they lack the ability to capture long-range dependencies. Transformer-based models \cite{liu2023itransformer,nietime} address this issue by employing self-attention to model global temporal interactions, achieving strong performance across benchmarks. Nonetheless, the quadratic complexity of attention and sensitivity to noise restrict their scalability and robustness in real-world scenarios. In parallel, lightweight MLP-based architectures \cite{tang2025unlocking,ma2024u} have recently emerged as promising alternatives, offering high efficiency but often struggling to represent heterogeneous and irregular patterns effectively.

\subsection{Patch-based Representation Learning}
To improve efficiency and capture fine-grained structures, patch-based strategies have gained increasing attention in time series modeling. Instead of processing sequences at raw temporal resolution, some methods \cite{wang2025towards,nietime,tang2025unlocking} divide time series into non-overlapping or partially overlapping patches, enabling models to operate on compact representations and reduce sequence length. While effective in long-horizon forecasting, these methods typically rely on continuous embeddings without explicit mechanisms to leverage recurring local shapes, which are prevalent in real-world time series. Vector quantization (VQ) \cite{van2017neural} provides a complementary perspective by discretizing local segments into a finite set of codewords, facilitating representation reuse and improving robustness, as extensively explored in domains such as vision and speech \cite{tian2024visual,wu2025janus}. Recent attempts \cite{shibo2025hdt,ansarichronos} to integrate quantization into time series tasks demonstrate its potential to capture recurring patterns efficiently. However, static or heuristic codebooks fail to adapt to real-world data dynamics.

Different from existing methods, ReCast innovatively propose a dual-path forecasting architecture with quantization, capturing both stable recurring shapes and irregular fluctuations. Besides, it introduces a reliability-aware updating incrementally refines codebook, ensuring robust adaptation to distribution shifts. 
\section{Methodology}
In this section, we present ReCast in detail, which has 3 modules: patch-wise quantization, dual-path forecasting, codebook construction and updating, as shown in Figure \ref{framework}. 
\subsection{Patch-wise Quantization}
Define the historical series as $\textbf{X}\in \mathbb{R}^{C\times L}=\{\textbf{x}_i\}_{i=1}^L$, and the ground truth future values as $\textbf{Y}\in \mathbb{R}^{C\times H}=\{\textbf{x}_i\}_{i=L+1}^{L+H}$. $L$ and $H$ are the length of the input and forecasting series, respectively. $C$ means the number of variables (or channels). $\textbf{x}_i$ is a vector of dimension $C$ at time step $i$. The goal of time series forecasting is to predict $\textbf{Y}$ based on observed $\textbf{X}$. ReCast first normalizes the input using instance normalization, which is $\textbf{X}=(\textbf{X}-\mu_{in})/\sqrt{\sigma_{in}^2+\varepsilon}$. $\mu_{in}$ and $\sigma_{in}$ denote the mean and variance of input, and $\varepsilon$ is a small constant added for numerical stability. The normalized $\textbf{X}$ is segmented into patches $\textbf{P}=\{\textbf{p}_{i}\}_{i=1}^{C\times N}$. $\textbf{p}_{i}\in \mathbb{R}^{L_p}$ is the $i$-th patch. $L_p$ is the patch length, and $N=\lceil L/L_p\rceil$. 

Each patch is subsequently quantized by assigning it to the nearest codeword in a learnable codebook $\textbf{S}=\{\textbf{s}_k\}_{k=1}^{K}$:
\begin{equation}
	\begin{aligned}
		&q_{i}=\textbf{S}(\tilde{\textbf{p}}_{i})=\mathop{\arg\min}\limits_{\textbf{s}_k\in \textbf{S}}||\tilde{\textbf{p}}_{i}-\textbf{s}_k||_2^2, \\
		&\tilde{\textbf{p}}_{i}=Dsamp(\textbf{p}_{i}),\ \ \textbf{s}_k,\tilde{\textbf{p}}_{i}\in \mathbb{R}^{L_p/2}
	\end{aligned}
	\label{quant}
\end{equation}
where $q_{i}\in\{1,\cdots,K\}$ is the discrete index associated with patch $\textbf{p}_{i}$. $K$ is the number of codewords. To reduce computational cost and suppress redundant local fluctuations, we apply downsampling $Dsamp(\cdot)$ on patches prior to quantization. This is supported by the well-established assumption in time series modeling that local patterns demonstrate invariance across scales and redundant morphology \cite{lu2022matrix}, which makes resolution reduction both meaningful and robust \cite{senin2013sax}. $\tilde{\textbf{p}}_{i}$ denotes the downsampled patch of $\textbf{p}_{i}$. This operation reduces the dimension of patches to $L_p/2$, significant savings in codebook matching, storage, and embedding projection. Additionally, it helps the codebook focus on salient structures, improving robustness and generalization to noisy or distribution shifts. The discrete embeddings for the full input series is organized as $\textbf{Q}_x=[\textbf{Q}_1;\cdots;\textbf{Q}_{C}]$, and $\textbf{Q}_i=\{q_{j}\}_{j=(i-1)\cdot N+1}^{i\cdot N}$ represents the discrete embedding of $i$-th variable. This discrete embeddings serves as the input to downstream forecasting modules.
\subsection{Dual-path Forecasting}
To simultaneously achieve computational efficiency and representational fidelity, ReCast adopts a dual-path forecasting architecture. This design decomposes the prediction task into two complementary paths, each responsible for capturing distinct aspects of temporal dynamics.
\subsubsection{Quantization path} 
To capturing the regular structures and modeling the evolution of local patterns, a lightweight multi-layer perceptron (MLP) $\mathcal{M}_\text{quant}$ is employed to forecast the discrete indices of future patches:
\begin{equation}
	\textbf{Q}_y=\mathcal{M}_\text{quant}(\textbf{Q}_x),
	\label{qpath}
\end{equation}
where $\textbf{Q}_y\in \mathbb{R}^{C\times N_y}$, and $N_y=\lceil H/L_p\rceil$. This path enables compact and efficient modeling of stable local patterns.
\subsubsection{Residual path}
While quantization promotes simplicity, it inevitably discards subtle variations. To mitigate this loss, ReCast introduces a residual correction branch. First, the input $\textbf{X}$ is approximately reconstructed from its quantized representation via codebook lookup:
\begin{equation}
	\begin{aligned}
		&\textbf{X}_{q}=Rec(\textbf{Q}_x|\textbf{S})=Rec(\textbf{Q}_1;\cdots;\textbf{Q}_C|\textbf{S}),\\ 
		&Rec(\textbf{Q}_i|\textbf{S})=Usamp([\textbf{s}_{q_{(i-1)\cdot N}}||\cdots||\textbf{s}_{q_{i\cdot N}}]),
	\end{aligned}
	\label{rec}
\end{equation}
where $\textbf{X}_{q}\in \mathbb{R}^{C\times L}$ denotes the approximate representation of $\textbf{X}$. $Rec(\textbf{Q}_i|\textbf{S})$ means reconstruction from discrete embedding $\textbf{Q}_i$ using the codebook $\textbf{S}$. $Usamp(\cdot)$ denotes the upsampling. $||$ denotes the concatenation. The residual component $\textbf{X}_r=\textbf{X}-\textbf{X}_q$ captures fine-scale discrepancies. A separate MLP forecaster $\mathcal{M}_\text{res}$ is trained to predict the residual signal for the future window:
\begin{equation}
	\textbf{Y}_r=\mathcal{M}_\text{res}(\textbf{X}_r), \ \ \textbf{Y}_r\in \mathbb{R}^{C\times H},
	\label{rpath}
\end{equation}

The final result combines both paths and is followed by instance denormalization to restore the original scale
\begin{equation}
	\begin{aligned}
		&\hat{\textbf{Y}}=\sigma_{in}(\textbf{Y}_q+\textbf{Y}_r)+\mu_{in}, \ \ \hat{\textbf{Y}}\in \mathbb{R}^{C\times H}, \\
		&\mathcal{L}_{pre}=||\hat{\textbf{Y}}-\textbf{Y}||_1,
	\end{aligned}
	\label{fpred}
\end{equation}
where $\textbf{Y}_q=Rec(\textbf{Q}_y|\textbf{S})$. To mitigate the distribution shift effect between the input $X$ and forecasting result, we use instance denormalization by $\sigma_{in}$ and $\mu_{in}$. We employ the $L_1$ Loss as the training objective to ensure robustness to outliers and stabilizes training.
\subsection{Codebook Construction and Updating}
The performance and robustness of ReCast are tightly coupled with the quality of its quantization codebook. Since real-world time series are often non-stationary and subject to distribution shifts, a static codebook is insufficient for capturing evolving local patterns. So, we adopt an incremental updating strategy for codebook construction, which allows the model to gradually refine its representation of local patterns based on data observed over time, as shown in stage 3 of Figure \ref{framework}. This approach can enable adaptation to evolving distributions, and avoid the instability and overfitting associated with outliers.
\subsubsection{Pseudo codebook construction}
At each epoch, we cluster the patches and obtain cluster centers. These centers are the representative local patterns that can be used to construct pseudo codebooks in the current epoch. The clustering objects are randomly sampled patches from the input. This random sampling reduces computational cost and prevents overfitting \cite{lu2022matrix,senin2013sax}. To ensure the efficiency, we express the energy function $\mathcal{L}_{c}$ of clustering in the form of matrix operation:
\begin{equation}
	\mathcal{L}_{c}=Tr((\tilde{\textbf{P}}^{t}-\textbf{M}\hat{\textbf{S}}^{t})^\top I(\tilde{\textbf{P}}^{t}-\textbf{M}\hat{\textbf{S}}^{t})),
	\label{kmeans}
\end{equation}
where $\hat{\textbf{S}}^{t}=\{\hat{\textbf{s}}^{t}_k\}_{k=1}^{K}$ and $\tilde{\textbf{P}}^{t}=\{\tilde{\textbf{p}}^{t}_i\}_{i=1}^{C\times N_p}$ denote the cluster center matrix and the sampled downsampled patches at $t$-th epoch, respectively. $\tilde{\textbf{p}}^{t}_i\in \mathbb{R}^{L_p/2}$ is the $i$-th patch of $\tilde{\textbf{P}}^{t}$, and $\hat{\textbf{s}}_k^t\in \mathbb{R}^{L_p/2}$ is the $k$-th cluster center of $\hat{\textbf{S}}^{t}$. $N_p$ is the number of sampled patches. $I$ is the weight matrix, here we take the identity matrix. $Tr(\cdot)$ is the trace of matrix. $\textbf{M}\in \mathbb{R}^{(C\times N_p) \times K}$ is the indicator matrix to indicate the membership of patches, which is a learnable binary matrix. $M_{i,j}=1$ means the patch $i$ belong to cluster $j$. The update function of cluster center is:
\begin{equation}
	\hat{\textbf{S}}^{t}=(\textbf{M}^\top I\tilde{\textbf{P}}^{t})/(\textbf{M}^\top I\textbf{M}),
	\label{center}
\end{equation}
the $\hat{\textbf{S}}^{t}$ reflects representative local patterns captured from the current training data distribution, which can serve as the pseudo codebook of $t$-th epoch.
\subsubsection{Incremental updating}
To ensure generalization to new patterns or distribution shifts, and avoid drastic changes of embeddings, we introduce a incremental updating strategy for codebook to balance adaptability and stability. At the first epoch, we initialize the codebook as $\textbf{S}^1=\hat{\textbf{S}}^{1}$. In subsequent epochs, the codebook is updated as:
\begin{equation}
	\textbf{S}^t = \textbf{S}^{t-1} + \frac{1}{t}(\textbf{W}^{t}\hat{\textbf{S}}^{t}-\textbf{S}^{t-1}), 
	\label{update}
\end{equation}
where $\textbf{S}^t$ denotes the codebook of $t$-th epoch. $\hat{\textbf{S}}^{t}$ is the pseudo codebook computed from the current epoch’s sampled patches via Equation \ref{center}. $\textbf{W}^{t}$ is a set of correction weights that adjust the influence of the current epoch’s pseudo codebook. $W_k^t$ is the weight for cluster center $\hat{s}^{t}_k$. Equation \ref{update} can ensure equitable contribution across epochs, while adaptively adjusting by $\textbf{W}^{t}$ (See \textbf{Appendix A.1} for complete proof). 
\subsubsection{Embedding regularization}
To promote better utilization of the embedding space and prevent codeword collapse, we introduce a limited separation loss that encourages diversity among the cluster centers:
\begin{equation}
	\mathcal{L}_{sep} = log\sum_{i,j=1}^{k}exp(-||\hat{\textbf{s}}^t_i-\hat{\textbf{s}}^t_j||_2^2)/\tau, 
	\label{seploss}
\end{equation}
where $\mathcal{L}_{sep}$ promotes the dispersion of embeddings in hidden space and prevents excessive expansion of the space by the temperature $\tau$. $\tau=||\hat{\textbf{S}}^t||_2^2$ ensures the embedding space size remains approximately consistent across each epoch. This loss penalizes excessive similarity among codewords, encouraging a well-distributed and expressive codebook.
\begin{figure}
	\centering
	\includegraphics[width=\linewidth]{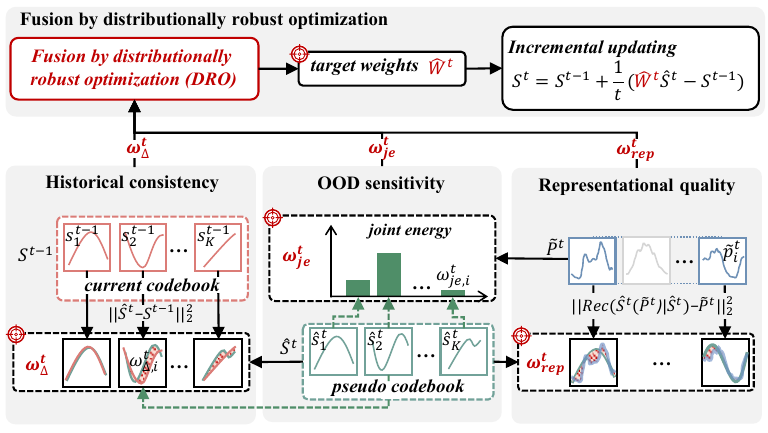}
	\caption{Illustration of reliability-aware scoring, showing three scoring factors and their fusion via distributionally robust optimization (DRO).}
	\label{scoring}
\end{figure}
\subsection{Reliability-aware Scoring}
As shown in Equation \ref{update}, $\textbf{W}^{t}$ can control the contribution of each pseudo codeword during updating. Rather than treating all cluster centers equally, ReCast introduces a reliability-aware scoring method that selectively integrates pseudo codewords based on their reliability to ensure robust and adaptive codebook updates. $\textbf{W}^{t}=\{w_k^t\}_{k=1}^{K}$ is computed by aggregating three complementary factors: $\textbf{w}_{rep}^t, \textbf{w}_{\Delta}^t, \textbf{w}_{je}^t\in \mathbb{R}^{K}$, and meets $\textbf{W}^{t} \propto \mathcal{M}_{fus}(\textbf{w}_{rep}^t, \textbf{w}_{\Delta}^t, \textbf{w}_{je}^t), \ \ \sum_{k=1}^{K}w^{t}_k=1$. $\mathcal{M}_{fus}$ is a fusion function. 
\subsubsection{Representational quality}
The $\textbf{w}_{rep}^t$ evaluates how well $\hat{\textbf{S}}_k^t$ represents its assigned patches, measured by the intra-cluster reconstruction error:
\begin{equation}
	w_{rep,k}^t = 1-\frac{exp({||\textbf{B}_k(Rec(\hat{\textbf{S}}^{t}(\tilde{\textbf{P}}^{t})|\hat{\textbf{S}}^{t})-\tilde{\textbf{P}}^{t})||_2^2})}{exp({||Rec(\hat{\textbf{S}}^{t}(\tilde{\textbf{P}}^{t})|\hat{\textbf{S}}^{t})-\tilde{\textbf{P}}^{t}||_2^2})+\varepsilon}, 
	\label{wrep}
\end{equation}
where $w_{rep,k}^t\in \textbf{w}_{rep}^t$ is the weight assigned to the $k$-th cluster center. $\textbf{B}_k$ is a binary matrix to mask values unrelated to the $k$-th cluster center. $Rec(\hat{\textbf{S}}^{t}(\tilde{\textbf{P}}^{t})|\hat{\textbf{S}}^{t})$ is the approximate representation reconstructed from discrete embeddings $\hat{\textbf{S}}^{t}(\tilde{\textbf{P}}^{t})$ using pseudo codebook $\hat{\textbf{S}}^{t}$. Higher value of $w_{rep,k}^t$ corresponds to better representational quality of the $k$-th cluster center, which has higher reliability.
\subsubsection{Historical consistency}
The $\textbf{w}_{\Delta}^t$ assesses the temporal stability of $\hat{\textbf{S}}_k^t$ by measuring its deviation from the corresponding codeword in the previous epoch:
\begin{equation}
	w_{\Delta,k}^t =\frac{exp({||\textbf{B}_k(\hat{\textbf{S}}^{t}-\textbf{S}^{t-1})||_2^2})}{exp({||\hat{\textbf{S}}^{t}-\textbf{S}^{t-1}||_2^2})+\varepsilon},
	\label{wdelta}
\end{equation}
where $w_{\Delta,k}^t\in \textbf{w}_{\Delta}^t$ is the weight assigned to the $k$-th cluster center. Higher value of $w_{\Delta,k}^t$ denotes the greater difference between $\hat{\textbf{s}}^{t}$ and $\textbf{s}^{t-1}$. Under the constraint of $\textbf{w}_{rep}^t$, this difference arises because $\textbf{S}^{t-1}$ lacks sufficient fitting capability for the newly input patches. So $\hat{\textbf{S}}^{t}$ should be given a greater weight to adjust the previous codebook, which is consistent with the expression of $\textbf{w}_{\Delta,k}^t$.
\subsubsection{OOD sensitivity}
The $\textbf{w}_{je}^t$ measures the OOD sensitivity of $\hat{\textbf{S}}^{t}$ by capturing potentially novel or rare patterns, estimated from assignment frequency and variance. The function is similar to joint-energy \cite{duvenaud2020your}:
\begin{equation}
	w_{je,k}^t =1-\frac{exp({\sum_{i=1}^{C\times N_p}|\tilde{\textbf{p}}^{t}_i-\hat{\textbf{s}}_k^t|})}{exp({\sum_{k=1}^{K}\sum_{i=1}^{C\times N_p}|\tilde{\textbf{p}}^{t}_i-\hat{\textbf{s}}_k^t|})+\varepsilon}, 
	\label{wje}
\end{equation}
where $w_{je,k}^t\in \textbf{w}_{je}^t$ is the weight assigned to the $k$-th cluster center. Higher value of $w_{je,k}^t$ indicates lower selection probabilities for the $k$-th cluster. By increasing its corresponding weight, we can prevent the embedding space of the codebook from collapsing into a few fixed codewords, and evaluate adaptability to OOD data \cite{duvenaud2020your}.
\subsubsection{Fusion by distributionally robust optimization}
In ReCast, each pseudo codeword is associated with three normalized reliability scores: $\textbf{w}_{rep}^t$, $\textbf{w}_{\Delta}^t$, and $\textbf{w}_{je}^t$. While these metrics are complementary, their relative importance may vary across epochs and data regimes. Directly assigning fixed weights can be suboptimal or unstable, especially when some metrics are noisy or biased due to transient data conditions \cite{duchi2019variance}. Thus, we formulate the fusion of reliability metrics as a distributionally robust optimization (DRO) problem \cite{qi2021online}. The goal is to obtain a conservative estimate of a codeword’s reliability by minimizing the expected reliability under the worst-case weighting distribution over the three metrics.

Formally, let the score vector for the $k$-th pseudo codeword at epoch $t$ be denoted as:
\begin{equation}
	\textbf{z}_k^t = [w_{rep,k}^t, w_{\Delta,k}^t, w_{je,k}^t] \in \mathbb{R}^3.
	\label{zkt}
\end{equation}
Instead of computing a simple average, we consider all possible distributions $\theta \in \Theta_3$ over the three scores, where $\Theta_3 = \{ \theta \in \mathbb{R}^3 \mid \sum_{i=1}^3 \theta_i = 1, \, \theta_i \geq 0 \}$. We then define the reliability score $\textbf{w}_k^t$ as the minimum expected value of $\textbf{z}_k^t$ under the worst-case distribution within a KL-divergence neighborhood around the uniform distribution $\mathbf{\textbf{u}} = [1/3, 1/3, 1/3]$:
\begin{equation}
	\hat{w}_k^t = \min_{\theta \in \mathcal{U}_\gamma} \langle \theta, \textbf{z}_k^t \rangle,
	\label{uni}
\end{equation}
where  $\mathcal{U}_\gamma = \left\{ \theta \in \Theta_3 \,\middle|\, \mathcal{D}_{KL}(\theta \,\|\, \mathbf{\textbf{u}}) \leq \gamma \right\}$. The parameter \(\gamma > 0\) determines the size of the uncertainty set: smaller values encourage near-uniform weighting, while larger values permit more skewed, adversarial distributions. This robust optimization problem has a closed-form solution (See \textbf{Appendix A.2} for complete proof):
\begin{equation}
	\hat{w}_k^t = -\gamma \cdot \log  \sum_{i=1}^{3} \exp( -\frac{\textbf{z}_{k,i}^t}{\gamma} ) .
	\label{closeform}
\end{equation}
The result is a soft-minimum over the scores, allowing the most reliable metric to dominate while softly discounting others. This formulation can be interpreted as an entropy-regularized minimization over reliability signals. 

By adopting this distributionally robust fusion scheme, ReCast is able to adaptively and conservatively score pseudo codewords, mitigating the impact of outliers or transient inconsistencies in individual metrics. This not only enhances the stability of the incremental codebook update but also improves the generalization of non-stationary time series.
\begin{table*}
	\centering
	\setlength{\tabcolsep}{1mm}{\begin{tabular}{c|cc|cc|cc|cc|cc|cc|cc|cc}
			\toprule
			Models&\multicolumn{2}{c|}{ReCast}&\multicolumn{2}{c|}{PatchMLP}&\multicolumn{2}{c|}{TQNet}&\multicolumn{2}{c|}{CycleNet}&\multicolumn{2}{c|}{iTransformer}&\multicolumn{2}{c|}{TimesNet}&\multicolumn{2}{c|}{PatchTST}&\multicolumn{2}{c}{Dlinear}\\
			\midrule
			Metric&MSE&MAE&MSE&MAE&MSE&MAE&MSE&MAE&MSE&MAE&MSE&MAE&MSE&MAE&MSE&MAE\\
			\midrule
			ETTm1&\textbf{0.371}&\textbf{0.379}&\underline{0.374}&\underline{0.382}&0.377&0.393&0.379&0.396&0.407&0.410&0.400&0.406&0.387&0.400&0.403&0.407\\
			ETTm2&\textbf{0.265}&\textbf{0.309}&0.269&\underline{0.311}&0.277&0.323&\underline{0.266}&0.314&0.286&0.327&0.291&0.333&0.281&0.326&0.350&0.401\\
			ETTh1&\textbf{0.437}&\textbf{0.428}&\underline{0.438}&\underline{0.429}&0.441&0.434&0.457&0.441&0.454&0.447&0.458&0.450&0.469&0.454&0.456&0.452\\
			ETTh2&\textbf{0.347}&\underline{0.385}&\underline{0.349}&\textbf{0.378}&0.378&0.402&0.388&0.409&0.383&0.407&0.414&0.427&0.387&0.407&0.559&0.515\\
			ECL&\textbf{0.163}&\textbf{0.257}&0.171&0.265&\underline{0.164}&\underline{0.259}&0.168&0.259&0.178&0.270&0.192&0.295&0.216&0.304&0.212&0.300\\
			Traffic&\underline{0.418}&\textbf{0.272}&\textbf{0.417}&\underline{0.273}&0.445&0.276&0.472&0.301&0.428&0.282&0.620&0.336&0.555&0.362&0.625&0.383\\
			Weather&\textbf{0.229}&\textbf{0.250}&\underline{0.231}&\underline{0.256}&0.242&0.269&0.243&0.271&0.258&0.279&0.259&0.287&0.259&0.281&0.265&0.317\\
			Solar&\underline{0.209}&\underline{0.260}&0.211&0.261&\textbf{0.198}&\textbf{0.256}&0.210&0.261&0.233&0.262&0.319&0.330&0.307&0.641&0.401&0.282\\
			\midrule
			$1^{st}$ Count&\multicolumn{2}{c|}{\textbf{12}}&\multicolumn{2}{c|}{2}&\multicolumn{2}{c|}{2}&\multicolumn{2}{c|}{0}&\multicolumn{2}{c|}{0}&\multicolumn{2}{c|}{0}&\multicolumn{2}{c|}{0}&\multicolumn{2}{c}{0} \\
			\bottomrule
	\end{tabular}}
	\caption{Comparison of forecasting performance. The best results are highlighted in \textbf{bold}, while the second-best results are \underline{underlined}. ReCast achieves the best performance in 12 out of 16 forecasting error metrics.}
	\label{main}
\end{table*}
\begin{table*}
	\centering
	\begin{tabular}{c|cc|cc|cc|cc|cc|cc}
		\toprule
		Setup&\multicolumn{2}{c|}{Original}&\multicolumn{2}{c|}{-Residual}&\multicolumn{2}{c|}{-Updating}&\multicolumn{2}{c|}{-Random}&\multicolumn{2}{c|}{-Scoring}&\multicolumn{2}{c}{-DRO}\\
		\midrule
		Metric&MSE&MAE&MSE&MAE&MSE&MAE&MSE&MAE&MSE&MAE&MSE&MAE\\
		\midrule
		ETTm1&\textbf{0.371}&\textbf{0.379}&0.377&0.395&0.400&0.402&0.377&0.396&0.385&0.399&0.375&0.385\\
		Traffic&\textbf{0.418}&\textbf{0.272}&0.435&0.281&0.553&0.332&0.427&0.285&0.441&0.285&0.424&0.281\\
		Weather&\textbf{0.229}&\textbf{0.250}&0.248&0.275&0.257&0.303&0.240&0.271&0.249&0.277&0.237&0.266\\
		\bottomrule
	\end{tabular}
	\caption{Ablation study of ReCast.}
	\label{ab1}
\end{table*}

Finally, the reliability score $\hat{\textbf{W}}^t=\{\hat{w}_k^t\}_{k=1}^{K}$ is used as a weighting coefficient to regulate the effect intensity of pseudo codewords in the codebook update. The Equation \ref{update} can been improved as:
\begin{equation}
	\textbf{S}^t = \textbf{S}^{t-1} + \frac{1}{t}(\hat{\textbf{W}}^{t}\hat{\textbf{S}}^{t}-\textbf{S}^{t-1}).
	\label{updatenew}
\end{equation}
\subsection{Learning Objective}
The final loss function is:
\begin{equation}
	\mathcal{L}= \mathcal{L}_{pre} + w_{sep}\mathcal{L}_{sep},
	\label{finalloss}
\end{equation}
where $w_{sep}$ is adjustment parameters. During the inference phase, the codebook remains fixed, and only Equation \ref{fpred} needs to be computed to efficiently obtain prediction results in a lightweight manner.
\section{Experiments}
\subsection{Datasets and Baselines}
We evaluate the proposed ReCast on 8 widely used real-world datasets: Electricity (ECL), Traffic, Weather, Solar \cite{liu2022scinet,Autoformer}, and 4 ETT datasets (ETTh1, ETTh2, ETTm1, ETTm2) \cite{zhou2021informer}. The full details of datasets presented in \textbf{Appendix B.1}. To evaluate the performance of ReCast, we compare it against 8 representative SOTA models from recent years: Transformer-based models: TQNet \cite{lintemporal}, iTransformer \cite{liu2023itransformer}, PatchTST \cite{nietime}; CNN-based model: TimeNet \cite{wu2023timesnet}; MLP-based models: PatchMLP \cite{kong2025unlocking}, CycleNet \cite{lin2024cyclenet}, DLinear \cite{Zeng2022AreTE}.
\subsection{Metrics and Implementation Details}
The models are evaluated based on both Mean Squared Error (MSE) and Mean Absolute Error (MAE). ReCast is implemented using Pytorch \cite{paszke2019pytorch} and trained on an Nvidia L40 GPU (48GB). The detailed implementations are described in \textbf{Appendix B.2}. The corresponding pseudocode of ReCast is provided in \textbf{Appendix B.3}.
\subsection{Main Results}
Table \ref{main} compares the forecasting performance of ReCast with baselines across 8 datasets, with lower MSE/MAE values indicating greater forecasting accuracy. ReCast achieves the best performance in 12 out of 16 forecasting error metrics, demonstrating overall SOTA accuracy. Full results are provided in \textbf{Appendix C.1}.

Notably, CNN-based models no longer retain a performance advantage due to their limited capacity in modeling long-range dependencies. Transformer-based models lies in modeling temporal contextual dependencies through attention mechanisms, which exhibit high sensitivity to noise. This inherent sensitivity limits their potential for further improving predictive performance. While they occasionally outperform simple MLP-based models, their performance is inconsistent, especially in noisy or irregular settings. Recent lightweight MLP-based models offer improved efficiency, but some of them often struggle to capture intricate inter-variable dependencies.

Moreover, channel-independent models (PatchTST and DLinear) often fail to realize their full potential, suggesting the irreplaceable role of inter-variable interactions. In contrast, ReCast employs a shared codebook across all variables, implicitly facilitating inter-variable interaction and thereby circumventing the performance limitations inherent in channel-independent architectures.
\subsection{Model Analysis}
\subsubsection{Ablation Study} Four variants are designed to assess the contributions of ReCast’s core components:`\textbf{-Residual}' disables the residual path, retaining only the quantization path; `\textbf{-Updating}'freezes the codebook, preventing incremental updates; `\textbf{-Random}'removes both downsampling during quantization and random sampling during codebook construction; `\textbf{-Scoring}' disables the reliability-aware fusion weights $\hat{W}_t$ in Equation~\ref{updatenew}, treating all pseudo codewords equally during codebook updates; `\textbf{-DRO}' uniformly weights the three scores. These variants can systematically evaluate the effects of dual-path architecture, robust enhancement operation, incremental updating, reliability-aware scoring, and DRO on model performance.
\begin{figure}
	\centering
	\includegraphics[width=\linewidth]{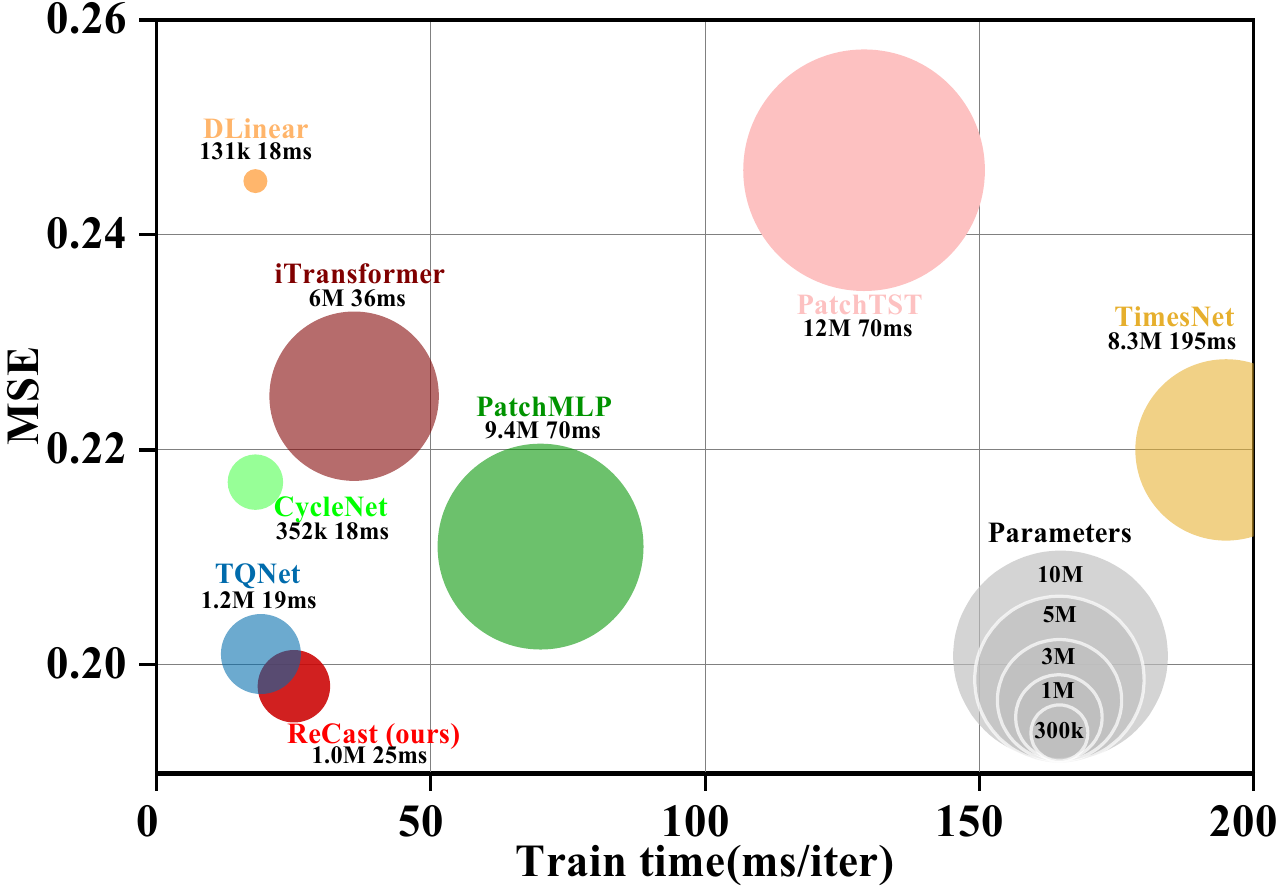}
	\caption{Computational efficiency of ReCast on the ECL dataset (horizon = 720).}
	\label{efficiency}
\end{figure}

The results of Table~\ref{ab1} lead to several key observations: 1) All ablated variants exhibit degraded performance relative to the full ReCast model, validating the effectiveness of each component. 2) The performance drop in `\textbf{-Residual}' highlights the critical role of the residual path in recovering fine-grained variations that are lost during quantization. 3)The performance deterioration in `\textbf{-Updating}' and `\textbf{-Scoring}' confirms that both dynamic codebook refinement and reliability-aware weighting are essential for capturing evolving local patterns and ensuring adaptability to distribution shifts. 4) The degradation observed in `\textbf{-Random}' underscores the importance of downsampling and random sampling for reducing overfitting and computational cost, while preserving performance. 5) The gap between `\textbf{-Scoring}' and `\textbf{-DRO}' reveals the importance of the DRO-based fusion strategy, which avoids over-reliance on any single score and enables robust reliability estimation.
\begin{figure*}
	\centering
	\includegraphics[width=\linewidth]{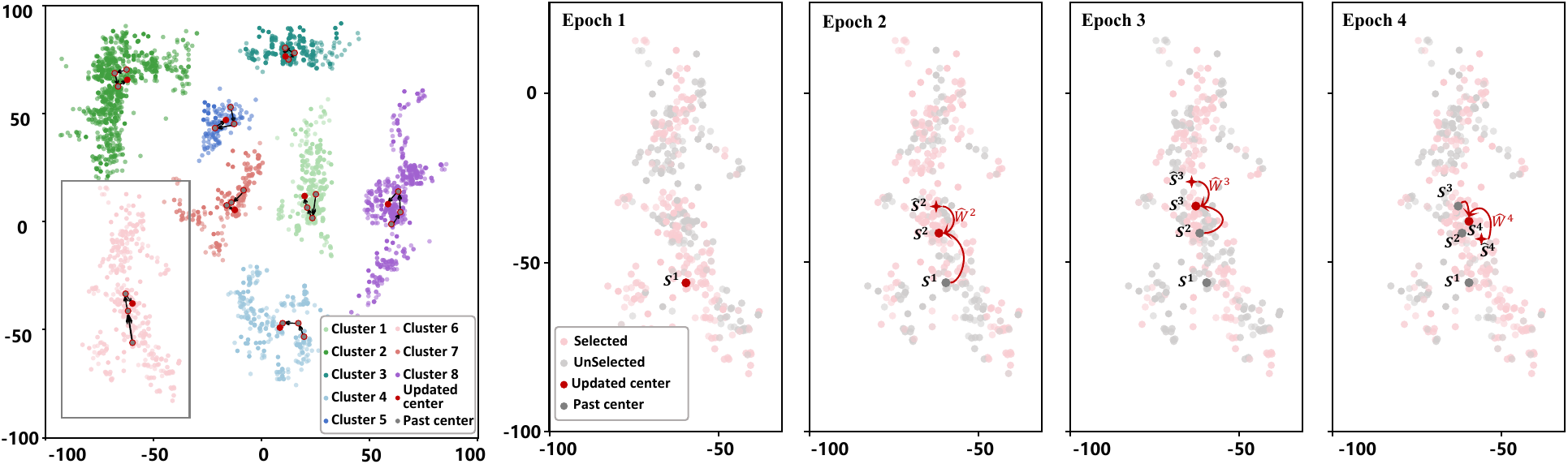}
	\caption{Visualization of codebook evolution and cluster assignments across epochs.}
	\label{vali}
\end{figure*}
\begin{table}
	\centering
	\setlength{\tabcolsep}{1.35pt}{\begin{tabular}{c|cc|cc|cc|cc}
			\toprule
			\multirow{2}{*}{Model}&\multicolumn{4}{c|}{iTransformer}&\multicolumn{4}{c}{TimesNet}\\
			&\multicolumn{2}{c|}{Original}&\multicolumn{2}{c|}{+ReCast}&\multicolumn{2}{c|}{Original}&\multicolumn{2}{c}{+ReCast}\\
			\midrule
			Metric&MSE&MAE&MSE&MAE&MSE&MAE&MSE&MAE\\
			\midrule
			ETTm1&0.407&0.410&\textbf{0.375}&\textbf{0.381}&0.400&0.406&\textbf{0.389}&\textbf{0.395}\\
			Traffic&0.428&0.282&\textbf{0.420}&\textbf{0.275}&0.620&0.336&\textbf{0.499}&\textbf{0.303}\\
			Weather&0.258&0.279&\textbf{0.231}&\textbf{0.259}&0.259&0.287&\textbf{0.245}&\textbf{0.272}\\
			\bottomrule
	\end{tabular}}
	\caption{Portability of ReCast across different backbones.}
	\label{ab2}
\end{table}
\subsubsection{Portability} To evaluate the portability of ReCast, we examine whether its dual-path forecasting architecture and reliability-aware codebook mechanism can generalize beyond the original MLP-based backbone. Specifically, we replace the MLP backbone with two widely used backbones: iTransformer, representative of Transformer-based methods, and TimesNet, representative of CNN-based methods. As reported on Table \ref{ab2}, integrating ReCast’s dual-path framework with either iTransformer or TimesNet improves forecasting performance. These results demonstrate that the proposed architecture is not tightly coupled with any specific backbone type and can be seamlessly adapted to a broad range of forecasting models, thereby confirming its strong portability and general applicability.
\begin{figure}
	\centering
	\includegraphics[width=0.49\linewidth]{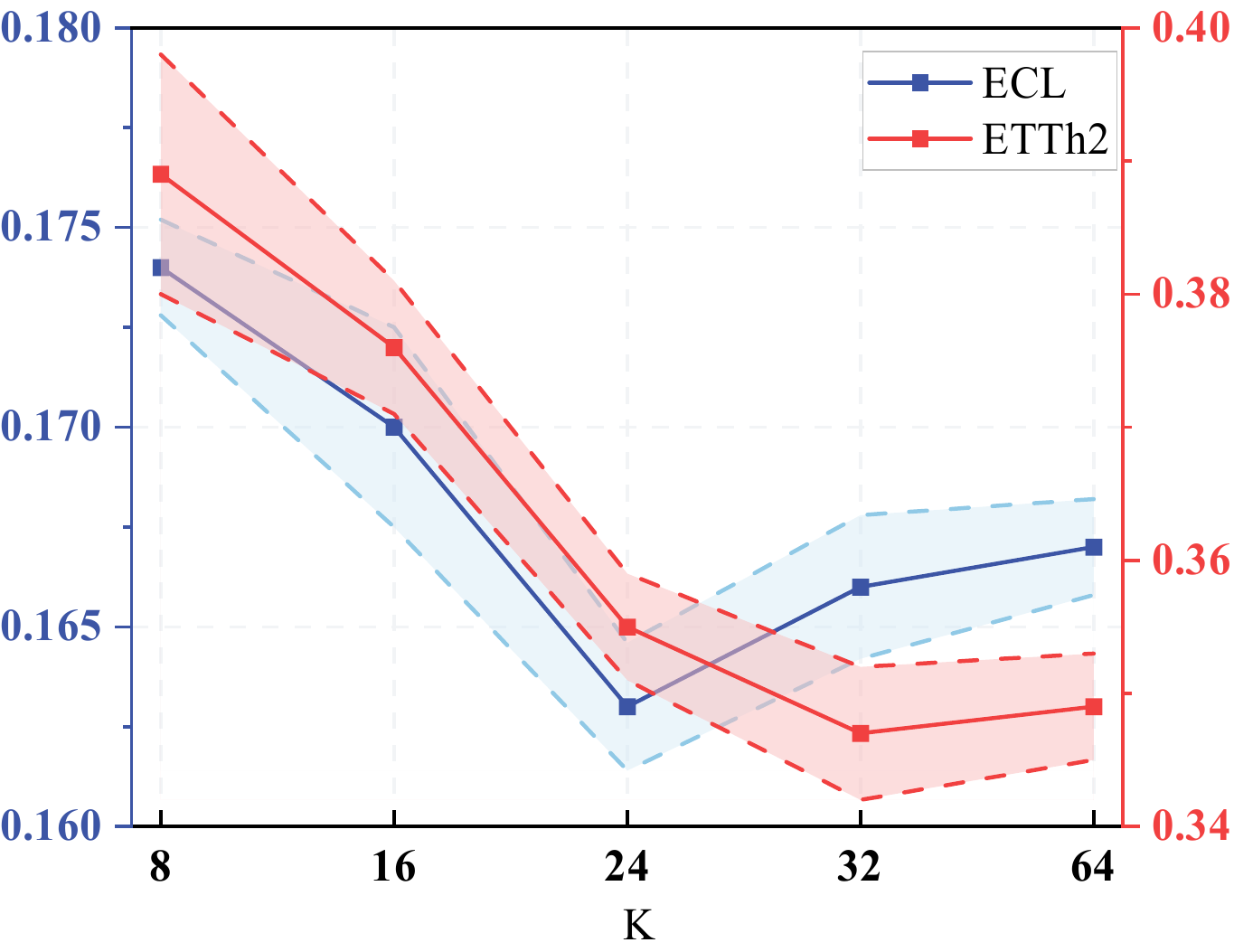}
	\includegraphics[width=0.49\linewidth]{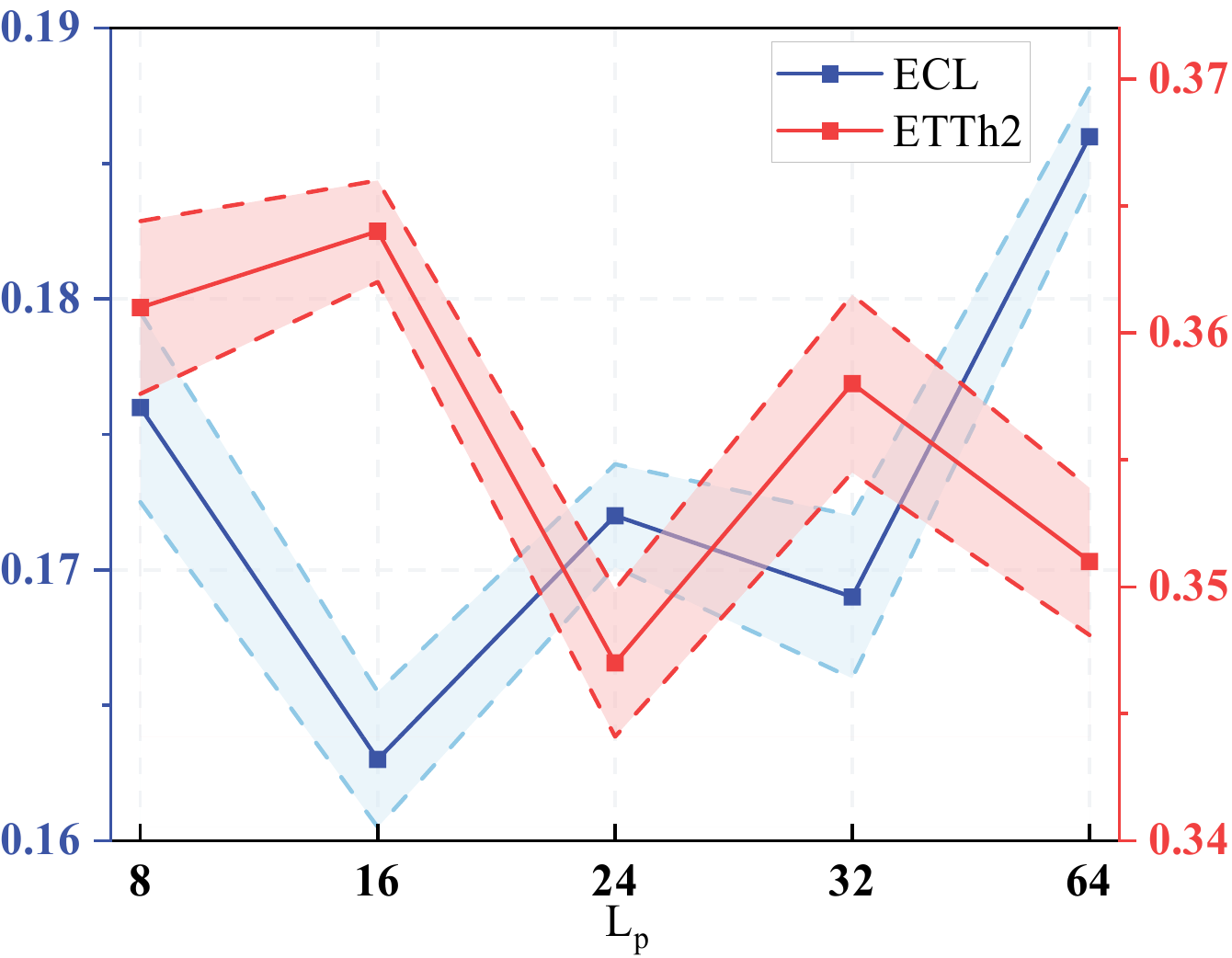}
	\caption{Performance comparison under varying hyper-parameters.}
	\label{para}
\end{figure}
\subsubsection{Efficiency} Benefiting from its lightweight dual-path architecture and a series of efficiency-oriented design choices, such as patch-wise quantization, residual correction, and selective sampling, ReCast achieves state-of-the-art forecasting accuracy while maintaining low computational overhead. As illustrated in Figure~\ref{efficiency}, ReCast consistently ranks among the top-performing models in terms of both parameter efficiency and training speed, without compromising predictive performance. These results highlight ReCast’s ability to strike a favorable balance between forecasting accuracy and computational complexity, making it well-suited for deployment in resource-constrained environments.
\subsubsection{Parameter sensitivity} Figure \ref{para} shows the performance under different hyperparameters (the number of clusters (codewords) $K$ and the patch length $L_p$).
\subsection{Visualization}
ReCast performs patch-wise clustering to generate discrete embeddings, its forecasting accuracy hinges on clustering quality and the representational capacity of the resulting cluster centers (codewords). To intuitively illustrate the codebook construction and update process, Figure~\ref{vali} presents qualitative visualizations. Representative examples of codewords are provided in the \textbf{Appendix C.2}.

The left side of Figure~\ref{vali} shows clustering results over 8 clusters and the evolution of cluster centers across epochs, where each color denotes a distinct cluster. Despite random sampling, cluster assignments remain stable and centers converge smoothly, demonstrating the robustness of the clustering. The right side of Figure~\ref{vali} illustrates the temporal dynamics of codebook updates. Taking epoch 2 as an example, the pseudo codebook $\hat{\textbf{S}}^2$ better fits the current data distribution than $\textbf{S}^1$, and the reliability-aware update assigns higher weight to $\hat{\textbf{S}}^2$, shifting $\textbf{S}^2$ closer to it. This confirms that the proposed reliability-aware update mechanism effectively balances adaptation and stability, supporting robust and generalizable forecasting.
\subsection{Limitations}
Despite its demonstrated accuracy and efficiency, ReCast presents a notable practical limitation: As shown in Figure \ref{para}, its performance is sensitive to the choice of $K$ and $L_p$. These parameters influence the trade-off between representational granularity and generalization capability to OOD patterns, yet are currently set empirically without adaptive or theoretical guidance. A promising direction is to scale ReCast to a pre-trained large language model with a richer codebook, diverse patch configurations, and heterogeneous time series pre-training, thereby improving robustness and reducing hyperparameter sensitivity.
\section{Conclusion}
In this work, we present ReCast, a novel codebook-assisted framework for reliable and efficient time series forecasting. Our dual-path architecture innovatively combines patch-wise quantization for capturing recurring local patterns with residual modeling for recovering irregular variations, achieving an optimal balance between lightweight design and forecasting accuracy. The proposed reliability-aware codebook update mechanism, supported by a reliability-aware scoring strategies, ensures robust adaptation to distribution shifts while maintaining stability. Extensive experiments across 8 real-world datasets demonstrate that ReCast outperforms SOTA baselines, achieving superior accuracy with significantly reduced computational complexity.

\section{Acknowledgments}
The authors appreciate the financial support by the National Natural Science Foundation of China (NSFC) Joint Fund with Zhejiang Integration of Informatization and Industrialization under Key Project (Grant Number U22A2033), the Postdoctoral Fellowship Program of CPSF under Grant Number GZC20251643, the NSFC under Grant Number 62576193.

\bibliography{aaai2026}

\end{document}

	%\onecolumn
	\section{A. Theoretical Proofs}
	\subsection{A.1 Proofs of Equation 8}
	\begin{equation}
		\begin{aligned}
			\mu_j^v &= \mu_{j-1}^v + \frac{1}{j}(w_{j}\hat{\mu}^v_{j}-\mu_{j-1}^v)=\frac{j-1}{j}\mu_{j-1}^v+\frac{w_{j}}{j}\hat{\mu}^v_{j}\\
			&=\frac{j-1}{j}(\frac{j-2}{j-1}\mu_{j-2}^v+\frac{w_{j-1}}{j-1}\hat{\mu}^v_{j-1})+\frac{w_{j}}{j}\hat{\mu}^v_{j}\\
			&=\frac{j-2}{j}\mu_{j-2}^v+\frac{w_{j-1}}{j-1}\hat{\mu}^v_{j-1}+\frac{w_{j}}{j}\hat{\mu}^v_{j}\\
			&\ \ \ \ \ \ \ \cdots\\
			&=\frac{1}{j}(w_{j_0}\hat{\mu}_{j_0}^v+w_{j_0+1}\hat{\mu}_{j_0+1}^v+\cdots+w_{j}\hat{\mu}^v_{j}).
			\label{proof}
		\end{aligned}
	\end{equation}
	
	We expand Equation (8) to analyze the contribution of pseudo-codebooks across epochs and the role of $W^t$:
	\begin{equation}
		\begin{aligned}
			\textbf{S}^t &= \textbf{S}^{t-1} + \frac{1}{t}(\textbf{W}^{t}\hat{\textbf{S}}^{t}-\textbf{S}^{t-1}) \\
			&=\frac{t-1}{t}\textbf{S}^{t-1}+\frac{\textbf{W}^{t}}{t}\hat{\textbf{S}}^{t}
		\end{aligned}
	\end{equation}
	By recursively applying this update rule, we obtain:
	\begin{equation}
		\begin{aligned}
			\textbf{S}^t &=\frac{t-1}{t}(\frac{t-2}{t-1}\textbf{S}^{t-2}+\frac{W^{t-1}}{t-1}\hat{\textbf{S}}^{t-1})+\frac{\textbf{W}^{t}}{t}\hat{\textbf{S}}^{t}\\
			&=\frac{t-2}{t}\textbf{S}^{t-2}+\frac{\textbf{W}^{t-1}}{t-1}\hat{\textbf{S}}^{t-1}+\frac{\textbf{W}^{t}}{t}\hat{\textbf{S}}^{t}\\
			&\cdots \\
			&=\frac{1}{t}(\textbf{W}^{1}\hat{\textbf{S}}^{1}+\textbf{W}^{2}\hat{\textbf{S}}^{2}+\cdots+\textbf{W}_{t}\hat{\textbf{S}}^{t})
		\end{aligned}
	\end{equation}
	This shows that every previous pseudo codebook contributes to \( \textbf{S}^t \) through a structurally uniform form, while the degree of influence is exponentially decayed unless \( \hat{\textbf{W}}^j \) is modulated.
	\subsection{A.2 Closed-form Derivation of Reliability Fusion via DRO}
	We present the derivation of the closed-form solution used in Equation~(14), which applies a distributionally robust optimization (DRO) framework to fuse multiple reliability scores under a KL-divergence constraint.
	\paragraph{Problem Formulation}
	We aim to compute the worst-case reliability score \( w_k^t \) for pseudo-codeword \( \hat{s}_k^t \) at timestep \( t \), defined by the following objective:
	\[
	w_k^t = \min_{\theta \in \mathcal{U}_\gamma} \langle \theta, \mathbf{z}_k^t \rangle,
	\]
	where \( \theta \in \mathbb{R}^3 \) is a probability vector over the three reliability metrics, \( \mathbf{z}_k^t = [w_{rep,k}^t, w_{\Delta,k}^t, w_{je,k}^t] \) contains the respective scores for representational quality, historical consistency, and OOD sensitivity, and \( \mathcal{U}_\gamma \) is a KL-divergence ball centered at the uniform distribution:
	\[
	\mathcal{U}_\gamma = \left\{ \theta \in \Theta_3 ~|~ D_{\mathrm{KL}}(\theta \| \mathbf{u}) \leq \gamma \right\},
	\]
	with \( \mathbf{u} = [1/3, 1/3, 1/3] \).
	
	\paragraph{Lagrangian Construction}
	Introducing a Lagrange multiplier \( \lambda \geq 0 \), we construct the Lagrangian:
	\[
	\mathcal{L}(\theta, \lambda) = \langle \theta, \mathbf{z}_k^t \rangle + \lambda \left( \sum_{i=1}^{3} \theta_i \log(3\theta_i) - \gamma \right).
	\]
	
	\paragraph{Optimality Condition}
	Taking the derivative w.r.t. \( \theta_i \) and setting it to zero yields:
	\[
	\frac{\partial \mathcal{L}}{\partial \theta_i} = z_{k,i}^t + \lambda ( \log(3\theta_i) + 1 ) = 0
	\]
	\[
	\Rightarrow \log(3\theta_i) = -\frac{z_{k,i}^t}{\lambda} - 1 \Rightarrow \theta_i \propto \exp\left( -\frac{z_{k,i}^t}{\lambda} \right)
	\]
	
	\paragraph{Normalization and Solution}
	Normalizing over \( i \in \{1,2,3\} \), we obtain the optimal weight vector:
	\[
	\theta_i^* = \frac{\exp(-z_{k,i}^t / \lambda)}{\sum_{j=1}^{3} \exp(-z_{k,j}^t / \lambda)}.
	\]
	Substituting \( \theta^* \) back into the original objective gives:
	\[
	\hat{w}_k^t = \sum_{i=1}^3 \theta_i^* z_{k,i}^t = -\lambda \log \sum_{i=1}^3 \exp\left( -\frac{z_{k,i}^t}{\lambda} \right).
	\]
	Letting \( \gamma = \lambda \), we arrive at the final closed-form:
	\[
	\hat{w}_k^t = -\gamma \cdot \log \sum_{i=1}^3 \exp\left( -\frac{z_{k,i}^t}{\gamma} \right).
	\]
	
	\paragraph{Interpretation}
	This expression is a softmin function that balances conservativeness and averaging:
	\begin{itemize}
		\item As \( \gamma \to 0 \), \( w_k^t \to \min_i z_{k,i}^t \)
		\item As \( \gamma \to \infty \), \( w_k^t \to \text{mean}(\mathbf{z}_k^t) \)
	\end{itemize}
	
	Such a form provides robustness against noisy or overly dominant scores, promoting conservative fusion under distributional uncertainty.

	\section{B. More Details of ReCast}
	\subsection{B.1 Datasets Details}
	Detailed information about datasets is provided in Table \ref{datasetdetail}.
	\begin{table}[h]
		\centering
		\resizebox{0.47\textwidth}{!}{\begin{tabular}{ccccc}
				\toprule
				Dataset      & Channels & Timesteps & Interval  & Domain \\ 
				\midrule
				ETTm1        &   7  & 57,600 & 15 mins  & Electricity   \\
				ETTm2        &   7  & 57,600 & 15 mins  & Electricity   \\
				ETTh1        &   7  & 14,400 & 1 hour   & Electricity   \\
				ETTh2        &   7  & 14,400 & 1 hour  & Electricity   \\
				ECL          & 321  & 26,304 & 1 hour   & Electricity   \\
				Traffic      & 862  & 17,544 & 1 hour   & Transportation\\
				Weather      &  21  & 52,696 & 10 mins  & Weather       \\
				Solar        & 137  & 52,560 & 10 mins  & Energy        \\
				\bottomrule
		\end{tabular}}
		\caption{Statistics of the benchmark datasets}
		\label{datasetdetail}
	\end{table}
	\subsection{B.2 Implementation Details}
	The data splits follow a 6:2:2 ratio for the ETT datasets and a 7:1:2 ratio for the remaining datasets. The input sequence is segmented into patches of length $L_p = 16$ and vector-quantised with a codebook of size $K \in \{8,\,16,\,24\}$. The quantization path implements as a single-layer MLP with hidden dimension $32$, and the residual path implements as a single-layer MLP with hidden dimension $512$. The batch size varies in [16,32,64] based on the dataset’s scale to maximize GPU utilization while avoiding out-of-memory errors. To improve generalisation, $50\%$ of the patches are randomly sampled during training. Training employs the Adam optimiser with an initial learning rate of $3\times10^{-4}$ under a cosine-annealing schedule. ReCast is trained for 30 epochs with early stopping based on a patience of 5 on the validation set. All experiments are conducted in \texttt{PyTorch} on a single NVIDIA L40 GPU (48 GB).
	\subsection{B.3 Pseudocode}
	\begin{algorithm}[h]
		\caption{Overall Pseudocode of \textbf{ReCast}}
		\KwIn{%
			historical series $\textbf{X}\!\in\!\mathbb{R}^{C\times L}$, forecast horizon $H$\\
			patch length $L_p$, codebook size $K$, training epochs $T$
		}
		\KwOut{%
			Forecasting series $\hat{\textbf{Y}}\in\mathbb{R}^{C\times H}$
		}
		
		\BlankLine
		$\textbf{X}\in \mathbb{R}^{C\times L}\leftarrow(\textbf{X}-\mu_{in})/\sqrt{\sigma_{in}^2+\varepsilon}$
		
		$\textbf{P}=\{\textbf{p}_{i}\}_{i=1}^{C\times N}$ $\leftarrow$ Split $\textbf{X}$ into length-$L_p$ patches
		
		\BlankLine
		\For{$t\leftarrow1$ \KwTo $T$}{
			$\tilde{\textbf{P}}^{t}=\{\tilde{\textbf{p}}^{t}_i\}_{i=1}^{C\times N_p}$ $\leftarrow$ Randomly downsample(\textbf{P})
			
			$\hat{\textbf{S}}^t=\{\hat{\textbf{s}}^{t}_k\}_{k=1}^{K}$ $\leftarrow$ Cluster($\tilde{\textbf{P}}^{t}$)
			
			\If {$t = 1$} {
				$\textbf{S}^1=\hat{\textbf{S}}^1$}
			\Else{
				$\hat{\textbf{W}}^t \in \mathbb{R}^{K}$ $\leftarrow$ Fuse Representational quality $\textbf{w}_{rep,k}$, Historical consistency $\textbf{w}_{\Delta,k}$, and OOD sensitivity $\textbf{w}_{je,k}$ via distributionally robust optimization
				
				$\displaystyle \textbf{S}^t \leftarrow \textbf{S}^{t-1}
				+ \frac1t\bigl(\hat{\textbf{W}}^t\,\hat{\textbf{S}}^t-\textbf{S}^{t-1}\bigr)$}

			$\textbf{Q}_x\in\mathbb{R}^{C\times N}$ $\leftarrow$ $\textbf{S}^t(\tilde{\textbf{P}}^{t})$
			
			$\textbf{Q}_y\in\mathbb{R}^{C\times N_y} \leftarrow MLP_{\text{quant}}(\textbf{Q}_x)$
			
			$\textbf{X}_r \in\mathbb{R}^{C\times L} \!\leftarrow\! \textbf{X}-\textbf{X}_q$, $\textbf{X}_q \!\leftarrow\! \operatorname{Rec}(\textbf{Q}_x\,|\,\textbf{S}^t)$
			
			$\textbf{Y}_r \leftarrow MLP_{\text{res}}(\textbf{X}_r)$
			
			$\textbf{Y}_q\in\mathbb{R}^{C\times H}\leftarrow\operatorname{Rec}(\textbf{Q}_y\,|\,\textbf{S}^t)$
			
			$\hat{\textbf{Y}}\in\mathbb{R}^{C\times H}\leftarrow\sigma_{\text{in}}(\textbf{Y}_q+\textbf{Y}_r)+\mu_{\text{in}}$
		}
	\end{algorithm}
	\section{C. More Results of ReCast}
	\subsection{C.1 Full Comparison Results}
	Table \ref{full} presents the full comparison results of ReCast against several baselines across 8 datasets. The results demonstrate that ReCast consistently achieves state-of-the-art forecasting performance under most experimental settings, underscoring the effectiveness of the proposed approach. For some baseline methods, only the average performance across all forecasting lengths is reported in the original papers, and their per-horizon results are unavailable. Therefore, we include only the average results for these methods in our comparison.
	\begin{table*}
		\centering
		\resizebox{\textwidth}{!}{\begin{tabular}{c|c|cc|cc|cc|cc|cc|cc|cc|cc}
				\toprule
				\multicolumn{2}{c|}{Models} & \multicolumn{2}{c|}{ReCast}  & \multicolumn{2}{c|}{PatchMLP} & \multicolumn{2}{c|}{TQNet} & \multicolumn{2}{c|}{CycleNet} & \multicolumn{2}{c|}{iTransformer} & \multicolumn{2}{c|}{TimesNet} & \multicolumn{2}{c|}{PatchTST} & \multicolumn{2}{c}{Dlinear}\\
				\midrule
				\multicolumn{2}{c|}{Metric} & MSE  & MAE & MSE & MAE & MSE & MAE & MSE & MAE & MSE & MAE & MSE & MAE & MSE & MAE & MSE & MAE \\
				\midrule
				\multirow{5}{*}{\rotatebox{90}{ETTm1}} & 96  & \textbf{0.308} & \textbf{0.345} & - & - & \underline{0.311} & \underline{0.353} & 0.319 & 0.360 & 0.334 & 0.368 & 0.338 & 0.375 & 0.329 & 0.367 & 0.345 & 0.372 \\ 
				~ & 192  & \textbf{0.352} & \textbf{0.360} & - & - & \underline{0.356} & \underline{0.378} & 0.360 & 0.381 & 0.377 & 0.391 & 0.374 & 0.387 & 0.367 & 0.385 & 0.380 & 0.389 \\ 
				~ & 336  & \textbf{0.385} & \textbf{0.381} & - & - & 0.390 & \underline{0.401} & \underline{0.389} & 0.403 & 0.426 & 0.420 & 0.410 & 0.411 & 0.399 & 0.410 & 0.413 & 0.413 \\ 
				~ & 720  & \textbf{0.439} & \textbf{0.431} & - & - & 0.452 & 0.440 & \underline{0.447} & 0.441 & 0.491 & 0.459 & 0.478 & 0.450 & 0.454 & \underline{0.439} & 0.474 & 0.453 \\ 
				~ & Avg  & \textbf{0.371} & \textbf{0.379} & \underline{0.374} & \underline{0.382} & 0.377 & 0.393 & 0.379 & 0.396 & 0.407 & 0.410 & 0.400 & 0.406 & 0.387 & 0.400 & 0.403 & 0.407 \\ \midrule 
				\multirow{5}{*}{\rotatebox{90}{ETTm2}} & 96  & \textbf{0.161} & \textbf{0.243} & - & - & 0.173 & 0.256 & \underline{0.163} & \underline{0.246} & 0.180 & 0.264 & 0.187 & 0.267 & 0.175 & 0.259 & 0.193 & 0.292 \\ 
				~ & 192  & \underline{0.231} & \textbf{0.287} & - & - & 0.238 & 0.298 & \textbf{0.229} & \underline{0.290} & 0.250 & 0.309 & 0.249 & 0.309 & 0.241 & 0.302 & 0.284 & 0.362 \\ 
				~ & 336  & \textbf{0.283} & \textbf{0.322} & - & - & 0.301 & 0.340 & \underline{0.284} & \underline{0.327} & 0.311 & 0.348 & 0.321 & 0.351 & 0.305 & 0.343 & 0.369 & 0.427 \\ 
				~ & 720  & \textbf{0.386} & \textbf{0.384} & - & - & 0.397 & 0.396 & \underline{0.389} & \underline{0.391} & 0.412 & 0.407 & 0.408 & 0.403 & 0.402 & 0.400 & 0.554 & 0.522 \\ 
				~ & Avg  & \textbf{0.265} & \textbf{0.309} & 0.269 & \underline{0.311} & 0.277 & 0.323 & \underline{0.266} & 0.314 & 0.288 & 0.332 & 0.291 & 0.333 & 0.281 & 0.326 & 0.350 & 0.401 \\ \midrule 
				\multirow{5}{*}{\rotatebox{90}{ETTh1}} & 96  & \textbf{0.368} & \textbf{0.387} & - & - & \underline{0.371} & \underline{0.393} & 0.375 & 0.395 & 0.386 & 0.405 & 0.384 & 0.402 & 0.414 & 0.419 & 0.386 & 0.400 \\ 
				~ & 192  & \textbf{0.426} & \textbf{0.423} & - & - & \underline{0.428} & \underline{0.426} & 0.436 & 0.428 & 0.441 & 0.436 & 0.436 & 0.429 & 0.460 & 0.445 & 0.437 & 0.432 \\ 
				~ & 336  & \textbf{0.470} & \textbf{0.440} & - & - & \underline{0.476} & \underline{0.446} & 0.496 & 0.455 & 0.487 & 0.458 & 0.491 & 0.469 & 0.501 & 0.466 & 0.481 & 0.459 \\ 
				~ & 720  & \textbf{0.485} & \textbf{0.462} & - & - & \underline{0.487} & \underline{0.470} & 0.520 & 0.484 & 0.503 & 0.491 & 0.521 & 0.500 & 0.500 & 0.488 & 0.519 & 0.516 \\ 
				~ & Avg  & \textbf{0.437} & \textbf{0.428} & \underline{0.438} & \underline{0.429} & 0.441 & 0.434 & 0.457 & 0.441 & 0.454 & 0.448 & 0.458 & 0.450 & 0.469 & 0.455 & 0.456 & 0.452 \\ \midrule 
				\multirow{5}{*}{\rotatebox{90}{ETTh2}} & 96  & \textbf{0.258} & \textbf{0.329} & - & - & \underline{0.295} & \underline{0.343} & 0.298 & 0.344 & 0.297 & 0.349 & 0.340 & 0.374 & 0.302 & 0.348 & 0.333 & 0.387 \\ 
				~ & 192  & \textbf{0.341} & \textbf{0.375} & - & - & \underline{0.367} & \underline{0.393} & 0.372 & 0.396 & 0.380 & 0.400 & 0.402 & 0.414 & 0.388 & 0.400 & 0.477 & 0.476 \\ 
				~ & 336  & \textbf{0.390} & \textbf{0.406} & - & - & \underline{0.417} & \underline{0.427} & 0.431 & 0.439 & 0.428 & 0.432 & 0.452 & 0.452 & 0.426 & 0.433 & 0.594 & 0.541 \\ 
				~ & 720  & \textbf{0.400} & \textbf{0.431} & - & - & 0.433 & 0.446 & 0.450 & 0.458 & \underline{0.427} & \underline{0.445} & 0.462 & 0.468 & 0.431 & 0.446 & 0.831 & 0.657 \\ 
				~ & ~  & \textbf{0.347} & \underline{0.385} & \underline{0.349} & \textbf{0.378} & 0.378 & 0.402 & 0.388 & 0.409 & 0.383 & 0.407 & 0.414 & 0.427 & 0.387 & 0.407 & 0.559 & 0.515 \\ \midrule 
				\multirow{5}{*}{\rotatebox{90}{ECL}} & 96  & \underline{0.135} & 0.234 & - & - & \textbf{0.134} & \textbf{0.229} & 0.136 & \underline{0.229} & 0.148 & 0.240 & 0.168 & 0.272 & 0.181 & 0.270 & 0.197 & 0.282 \\ 
				~ & 192  & 0.155 & \textbf{0.244} & - & - & \underline{0.154} & 0.247 & \textbf{0.152} & \underline{0.244} & 0.162 & 0.253 & 0.184 & 0.289 & 0.188 & 0.274 & 0.196 & 0.285 \\ 
				~ & 336  & \textbf{0.163} & \textbf{0.261} & - & - & \underline{0.169} & \underline{0.264} & 0.170 & 0.264 & 0.178 & 0.269 & 0.198 & 0.300 & 0.204 & 0.293 & 0.209 & 0.301 \\ 
				~ & 720  & \textbf{0.200} & \textbf{0.290} & - & - & \underline{0.201} & \underline{0.294} & 0.212 & 0.299 & 0.225 & 0.317 & 0.220 & 0.320 & 0.246 & 0.324 & 0.245 & 0.333 \\ 
				~ & Avg  & \textbf{0.163} & \textbf{0.257} & 0.171 & 0.265 & \underline{0.164} & \underline{0.259} & 0.168 & 0.259 & 0.178 & 0.270 & 0.193 & 0.295 & 0.205 & 0.290 & 0.212 & 0.300 \\ \midrule 
				\multirow{5}{*}{\rotatebox{90}{Traffic}} & 96  & \textbf{0.382} & \textbf{0.257} & - & - & 0.413 & \underline{0.261} & 0.458 & 0.296 & \underline{0.395} & 0.268 & 0.593 & 0.321 & 0.462 & 0.290 & 0.650 & 0.396 \\ 
				~ & 192  & \textbf{0.406} & \textbf{0.268} & - & - & 0.432 & \underline{0.271} & 0.457 & 0.294 & \underline{0.417} & 0.276 & 0.617 & 0.336 & 0.466 & 0.290 & 0.598 & 0.370 \\ 
				~ & 336  & \textbf{0.423} & \textbf{0.273} & - & - & 0.450 & \underline{0.277} & 0.470 & 0.299 & \underline{0.433} & 0.283 & 0.629 & 0.336 & 0.482 & 0.300 & 0.605 & 0.373 \\ 
				~ & 720  & \textbf{0.459} & \textbf{0.291} & - & - & 0.486 & \underline{0.295} & 0.502 & 0.314 & \underline{0.467} & 0.302 & 0.640 & 0.350 & 0.514 & 0.320 & 0.645 & 0.394 \\ 
				~ & Avg  & \underline{0.418} & \textbf{0.272} & \textbf{0.417} & \underline{0.273} & 0.445 & 0.276 & 0.472 & 0.301 & 0.428 & 0.282 & 0.620 & 0.336 & 0.481 & 0.300 & 0.625 & 0.383 \\ \midrule 
				\multirow{5}{*}{\rotatebox{90}{Weather}} & 96  & \textbf{0.141} & \textbf{0.181} & - & - & \underline{0.157} & \underline{0.200} & 0.158 & 0.203 & 0.174 & 0.214 & 0.172 & 0.220 & 0.177 & 0.210 & 0.196 & 0.255 \\ 
				~ & 192  & \textbf{0.189} & \textbf{0.230} & - & - & \underline{0.206} & \underline{0.245} & 0.207 & 0.247 & 0.221 & 0.254 & 0.219 & 0.261 & 0.225 & 0.250 & 0.237 & 0.296 \\ 
				~ & 336  & \textbf{0.253} & \textbf{0.267} & - & - & \underline{0.262} & \underline{0.287} & 0.262 & 0.289 & 0.278 & 0.296 & 0.280 & 0.306 & 0.278 & 0.290 & 0.283 & 0.335 \\ 
				~ & 720  & \textbf{0.333} & \textbf{0.323} & - & - & \underline{0.344} & 0.342 & 0.344 & 0.344 & 0.358 & 0.349 & 0.365 & 0.359 & 0.354 & \underline{0.340} & 0.345 & 0.381 \\ 
				~ & Avg  & \textbf{0.229} & \textbf{0.250} & \underline{0.231} & \underline{0.256} & 0.242 & 0.269 & 0.243 & 0.271 & 0.258 & 0.278 & 0.259 & 0.287 & 0.259 & 0.273 & 0.265 & 0.317 \\ \midrule 
				\multirow{5}{*}{\rotatebox{90}{Solar}} & 96  & \underline{0.181} & \underline{0.237} & - & - & \textbf{0.173} & \textbf{0.233} & 0.190 & 0.247 & 0.203 & 0.237 & 0.250 & 0.292 & 0.234 & 0.286 & 0.290 & 0.378 \\ 
				~ & 192  & 0.215 & \underline{0.261} & - & - & \textbf{0.199} & \textbf{0.257} & \underline{0.210} & 0.266 & 0.233 & 0.261 & 0.296 & 0.318 & 0.267 & 0.310 & 0.320 & 0.398 \\ 
				~ & 336  & 0.221 & 0.273 & - & - & \textbf{0.211} & \textbf{0.263} & \underline{0.217} & \underline{0.266} & 0.248 & 0.273 & 0.319 & 0.330 & 0.290 & 0.315 & 0.353 & 0.415 \\ 
				~ & 720  & \underline{0.217} & \underline{0.270} & - & - & \textbf{0.209} & 0.270 & 0.223 & \textbf{0.266} & 0.249 & 0.275 & 0.338 & 0.337 & 0.289 & 0.317 & 0.356 & 0.413 \\ 
				~ & Avg  & \underline{0.209} & \underline{0.260} & 0.211 & 0.261 & \textbf{0.198} & \textbf{0.256} & 0.210 & 0.261 & 0.233 & 0.262 & 0.301 & 0.319 & 0.270 & 0.307 & 0.330 & 0.401 \\ 			\bottomrule 
		\end{tabular}}
		\caption{Full time series forecasting results for all prediction horizons $H\in \{96,192,336,720\}$. The look-back length $L=96$, and the reproduced baseline results are sourced from TQNet. The best results are highlighted in \textbf{bold}, while the second-best results are \underline{underlined}.}
		\label{full}
	\end{table*}
	\subsection{C.2 Examples of codewords}
	To provide a clearer understanding of the learned discrete embeddings, we present representative examples of codewords under different configurations. Specifically, we consider two settings:
	\begin{itemize}
		\item \textbf{Patch length $L_p = 16$, with $K = 8$ clusters}: The time series is segmented into patches of length 16, and each patch is quantized into one of 8 codewords. Representative examples from each cluster are visualized to illustrate the diversity and semantics captured by the codebook.
		
		\item \textbf{Patch length $L_p = 24$, with $K = 16$ clusters}: The time series is divided into longer patches of length 24, and quantized into 16 codewords. We display representative instances for each codeword to demonstrate the increased granularity and expressiveness provided by the larger codebook.
	\end{itemize}
	
\begin{figure*}
		\centering
		\includegraphics[width=\linewidth]{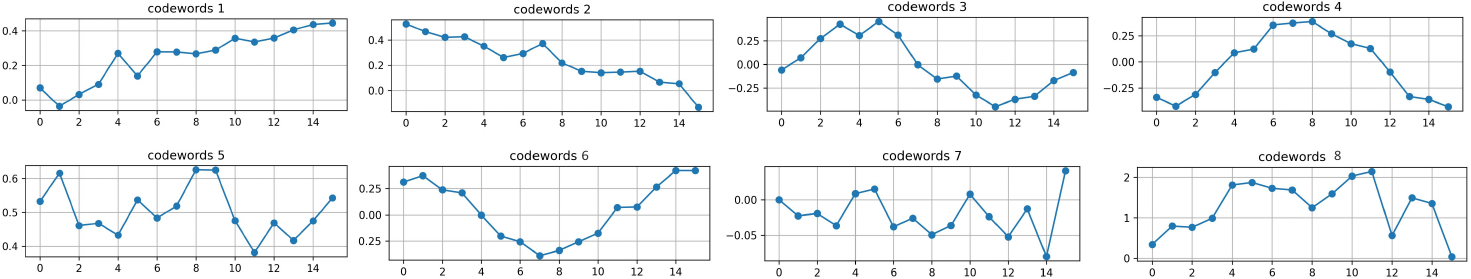}
		\caption{Representative examples of the learned codewords under the setting of window length $L_p = 16$ and codebook size $K = 8$.}
		\label{16}
	\end{figure*}

\begin{figure*}
		\centering
		\includegraphics[width=\linewidth]{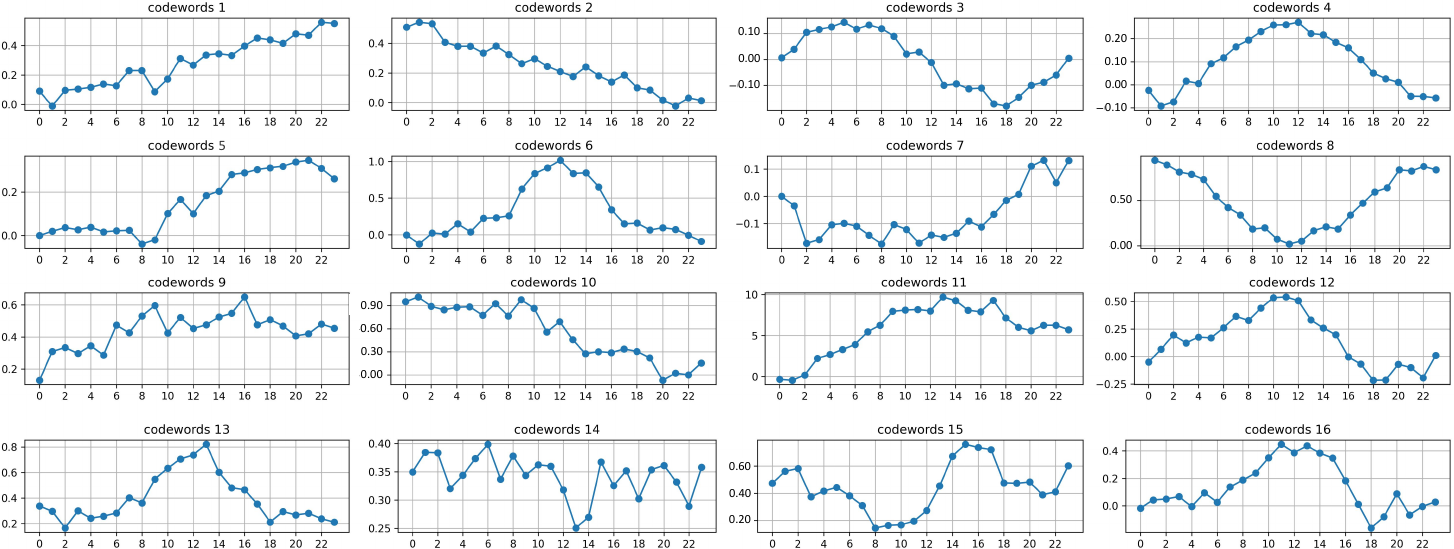}
		\caption{Representative examples of the learned codewords under the setting of window length $L_p = 24$ and codebook size $K = 16$.Compared to the smaller configuration, the increased window length and codebook size lead to more expressive and fine-grained representations.}
		\label{24}
	\end{figure*}

	The corresponding visualizations are shown in Figure~\ref{16} and Figure~\ref{24}, respectively. These visualizations demonstrate the structure and interpretability of the learned codewords, and how varying the patch length and codebook size influences the representation capacity.
	
	\newpage
